\documentclass[11pt,a4paper]{article}
\usepackage[hyperref]{emnlp2020}
\usepackage{times}
\usepackage{latexsym}

\usepackage{xcolor}
\usepackage{graphicx}  
\usepackage{subcaption} 
\usepackage{booktabs} 
\usepackage{multirow}
\usepackage{amsmath} 
\usepackage{amssymb}
\usepackage{mathtools}
\usepackage{bbm}
\usepackage{enumitem}
\usepackage[ruled,vlined,linesnumbered]{algorithm2e} 
\usepackage[normalem]{ulem}
\usepackage{xcolor} 

\newcommand{\comm}[1]{}

\usepackage{microtype}

\aclfinalcopy 
\def\aclpaperid{1100} 

\title{Iterative Domain-Repaired Back-Translation}

\author{
Hao-Ran Wei, Zhirui Zhang, Boxing Chen \and Weihua Luo \\
Machine Intelligence Technology Lab \\ 
Alibaba Group \\ 
Hangzhou, China \\
{\{funan.whr, zhirui.zzr, boxing.cbx, weihua.luowh\}@alibaba-inc.com}
}

\date{}

\begin{document}
\maketitle
\begin{abstract}

In this paper, we focus on the domain-specific translation with low resources, where in-domain parallel corpora are scarce or nonexistent.
One common and effective strategy for this case is exploiting in-domain monolingual data with the back-translation method.
However, the synthetic parallel data is very noisy because they are generated by imperfect out-of-domain systems, resulting in the poor performance of domain adaptation.
To address this issue, we propose a novel iterative domain-repaired back-translation framework, which introduces the Domain-Repair (DR) model to refine translations in synthetic bilingual data. 
To this end, we construct corresponding data for the DR model training by round-trip translating the monolingual sentences, and then design the unified training framework to optimize paired DR and NMT models jointly.
Experiments on adapting NMT models between specific domains and from the general domain to specific domains demonstrate the effectiveness of our proposed approach, achieving 15.79 and 4.47 BLEU improvements on average over unadapted models and back-translation.\footnote{Our code is released in \url{https://github.com/whr94621/Iterative-Domain-Repaired-Back-Translation}}
 
\end{abstract}

\section{Introduction}

Neural Machine Translation (NMT) has achieved impressive performance when large amounts of parallel sentences are available~\cite{Wu2016GooglesNM,vaswani2017attention,Hassan2018AchievingHP}. 
However, some previous works have shown that NMT models perform poorly in specific domains, especially when they are trained on the corpora from very distinct domains~\cite{Koehn2017SixCF,Chu2018ASO}.
The fine-tuning method~\cite{luong2015stanford} is a popular way to mitigate the effect of domain drift. However, it is not realistic to collect large amounts of high-quality parallel data in every domain we are interested in.
Since monolingual in-domain data are usually abundant and easy to obtain, it is essential to explore the unsupervised domain adaptation scenario that utilizes large amounts of out-of-domain bilingual data and in-domain monolingual data.

One straightforward and effective solution for unsupervised domain adaptation is to build in-domain synthetic parallel data, including copying monolingual target sentences to the source side~\cite{Currey2017CopiedMD} or back-translation of in-domain monolingual target sentences~\cite{sennrich2016improving,Dou2019UnsupervisedDA}.
Although the back-translation approach has proven the superior effectiveness in exploiting monolingual data, directly applying this method in this scenario brings low-quality in-domain synthetic data.
Table \ref{tab:mistakes-of-bt} gives two incorrect translation sentences generated by back-translation method.
The main reason for this situation is that the synthetic parallel data is built by imperfect out-of-domain NMT systems, which leads to inappropriate word expressions or wrong translations.
Fine-tuning on such synthetic data is very likely to hurt the performance of domain adaptation.

In this paper, we extend back-translation by a Domain-Repair (DR) model to explicitly remedy this issue.
Specifically, the DR model is designed to re-generate in-domain source sentences given the synthetic data.
In this way, the pseudo parallel data's source side can be re-written with the in-domain style, and some wrong translations are fixed.
To optimize the DR model, we use the round-trip translation of monolingual source sentences to construct the corresponding training data.

\begin{table}[t]
  \footnotesize
  \centering
    \begin{tabular}{lp{19.5em}}
    \toprule
    \textbf{SRC:} &  eine Gewichtszunahme wurde nach \textit{Markteinführung} bei Patienten berichtet , denen ABILIFY verschrieben wurde . \\
    \textbf{REF:} & weight gain has been reported \textcolor{red}{\textbf{\uline{post-marketing}}} among patients prescribed ABILIFY . \\
    \textbf{MT:} & a weight gain has been reported \textcolor{blue}{\textbf{\uwave{after market introduction}}} in patients who have been prescribed ABILIFY . \\
    \midrule
    \textbf{SRC:} & es werden möglicherweise nicht alle Packungsgrößen \textit{in den Verkehr gebracht} . \\
    \textbf{REF:} & not all pack sizes may be  \textcolor{red}{\textbf{\uline{marketed}}}. \\
    \textbf{MT:} & it may not all pack sizes may be \textcolor{blue}{\textbf{\uwave{added to the pack}}} . \\
    \bottomrule
    \end{tabular}%
    \vspace{-5pt}
    \caption{Two incorrect medical translations caused by the law-domain NMT model in German-English multi-domain datasets~\cite{TIEDEMANN12.463}, in which ``\textit{Markteinführung}'' and ``\textit{in den Verkehr gebracht}'' are translated to ``after market introduction'' and ``added to the pack'' respectively.}
  \label{tab:mistakes-of-bt}%
\end{table}%

\begin{figure}[t]
  \centering
  \includegraphics[width=0.50\textwidth]{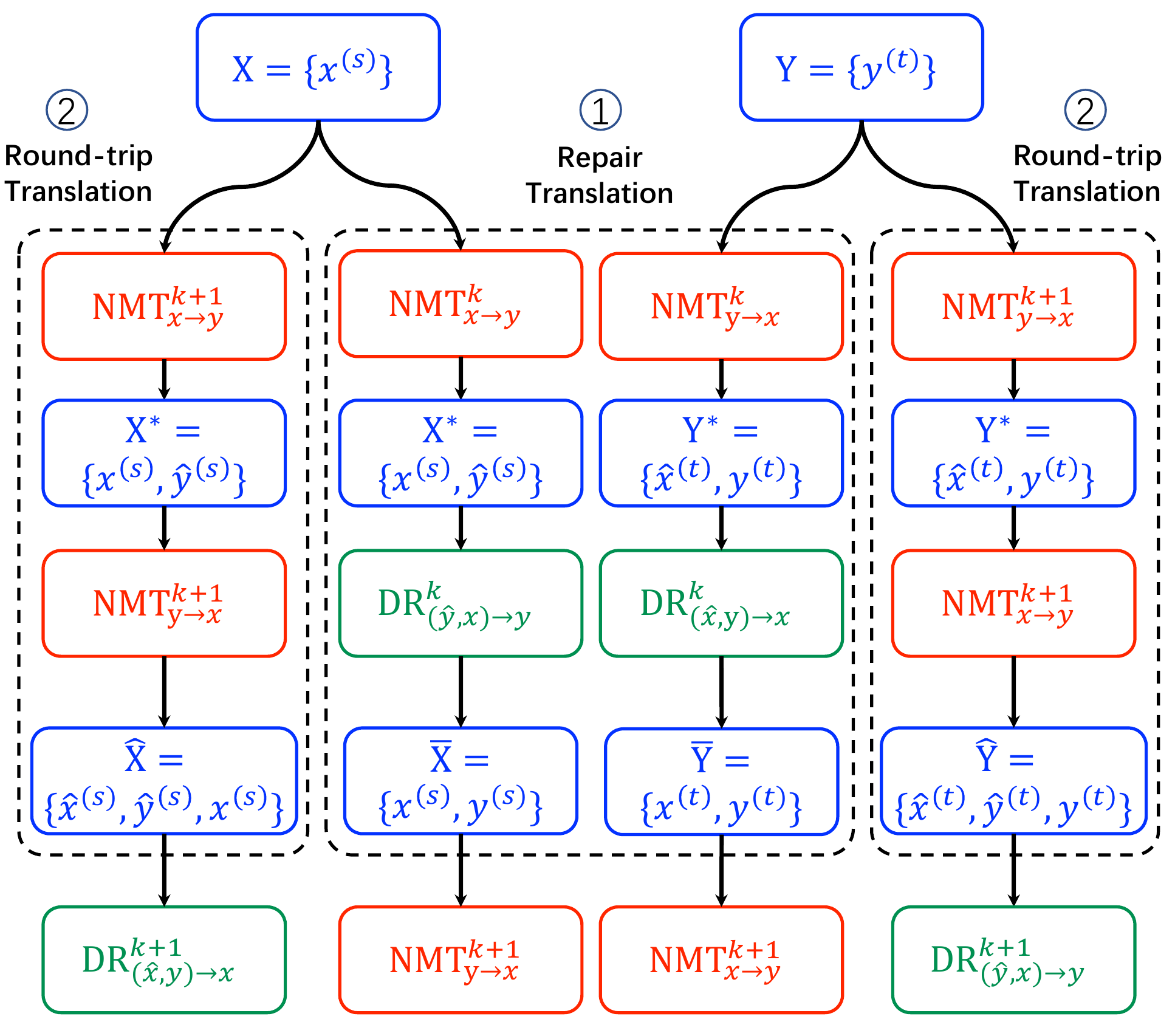}
\caption{The training process of the iterative domain-repaired back-translation (iter-DRBT) framework at epoch $k$, where $x$ and $y$ represent the source and target sentences respectively, $\widehat{x}$ and $\widehat{y}$ denote the translation generated by NMT models. The whole framework consists of translation repair and round-trip translation procedures, which are used to generate corresponding training data for NMT and DR models respectively. }
\label{fig:iter-DRBT-overview}
\end{figure}

Since source monolingual data is involved, it is natural to extend the back-translation method to bidirectional setting~\cite{zhang2018joint}, which jointly optimizes source-to-target and target-to-source NMT models. 
Based on this setting, we propose the iterative domain-repaired back-translation (iter-DRBT) framework to fully exploit both source and target in-domain monolingual data.
The whole framework starts with pre-trained out-of-domain bidirectional NMT models, and then these models are adopted to perform round-trip translation on monolingual data to obtain initial bidirectional DR models.
Next, as illustrated in Figure \ref{fig:iter-DRBT-overview}, we design a unified training algorithm consisting of translation repair and round-trip translation procedures to jointly update DR and NMT models. More particularly, 
in the translation repair stage, the back-translated synthetic data can be well re-written as in-domain sentences by the well-trained DR models to further improve NMT models.
Then enhanced NMT models run the round-trip translation on monolingual data to build domain-mapping data, which helps DR models better identify mistakes made by the latest NMT models.
This training process is iteratively carried out to make full use of the advantage of DR models to improve NMT models.

We evaluate our proposed method on German-English multi-domain datasets \cite{TIEDEMANN12.463}.
Experimental results on adapting NMT models between specific domains and from the general domain to specific domains show that our proposed method obtains 15.79 and 4.47 BLEU improvements on average over unadapted models and back-translation, respectively.
Further analysis demonstrates the ability of DR models to repair the synthetic parallel data.

\section{Related Work}

Since in-domain parallel corpora are usually hard to obtain, many studies attempt to improve the performance of NMT models without any in-domain parallel sentences. 
One research line is to extract pseudo in-domain data from large amounts of out-domain parallel data. 
\citet{Biici2011InstanceSF} use an in-domain held-out set to obtain parallel sentences from out-domain parallel sentences by computing n-gram overlaps. 
Instead, \citet{Moore2010IntelligentSO}, \citet{Axelrod2011DomainAV} and \citet{Duh2013AdaptationDS} use LMs score to select data similar to in-domain text.
Recently, \citet{Chen2017CostWF} train a domain classifier to weight the out-domain training samples. 
There are also work on adaptation via retrieving sentences or n-grams in the training data similar to the test set \cite{Farajian2017MultiDomainNM, Bapna2019NonParametricAF}. 
However, these methods cannot always guarantee to find domain-specific samples from out-domain data.

Another research direction is to exploit plenty of in-domain monolingual data, e.g., integrating a language model during decoding~\cite{Glehre2015OnUM}, copy method~\cite{Currey2017CopiedMD}, back-translation~\cite{sennrich2016improving} or obtaining domain-aware feature embedding via an auxiliary language modeling~\cite{Dou2019UnsupervisedDA}.
Among these approaches, back-translation is a widely used and effective method in exploiting monolingual data.
Our proposed method is also based on back-translation and makes the most of it by improving the data quality with the DR model. 

The methods of exploiting monolingual data in NMT can be naturally applied in unsupervised domain adaptation. 
Some studies are working on exploiting source-side monolingual data by self-training~\cite{zhang2016exploiting,chinea-rios-etal-2017-adapting} or pre-training~\cite{yang2019towards, weng2019acquiring,Ji2020CrosslingualPB}, and leveraging both source and target monolingual data simultaneously by semi-supervised learning~\cite{Cheng2016SemiSupervisedLF}, dual learning~\cite{he2016dual} and joint training \cite{zhang2018joint,hoang-etal-2018-iterative}.
Our method utilizes both source and target data as well, with different that we use monolingual data to train bidirectional DR models, and then these models are used to fix pseudo data.

As back-translation is widely considered more effective than the self-training method, several works find that performance of back-translation degrades due to the less rich translation or domain mismatch at the source side of the synthetic data \cite{edunov2018understanding,caswell-etal-2019-tagged}. 
\citet{edunov2018understanding} attempt to use sampling instead of maximum a-posterior when decoding with the reverse direction model. 
\citet{imamura-etal-2018-enhancement} add noises to the results of beam search. 
\citet{caswell-etal-2019-tagged} propose to add a tag token at the source side of the synthetic data. 
Unlike their methods, our method leverages the DR model to re-generate the source side of the synthetic data, which can also increase translation diversity and mitigate the effect of different domains.

\section{Iterative Domain-Repaired Back-Translation}

In this section, we first illustrate the overview of iter-DRBT framework, then describe the architecture of DR model and the joint training strategy.

\subsection{Overview}

Suppose that we have non-parallel in-domain monolingual sentences $X=\{x^{(s)}\}$ and $Y=\{y^{(t)}\}$ in two languages respectively, as well as two pre-trained out-of-domain translation models $\text{NMT}^{0}_{x \to y}$ and $\text{NMT}^{0}_{y \to x}$, where $x$ and $y$ denote the source and target sentences respectively. 
The purpose of unsupervised domain adaptation is to train in-domain models $\text{NMT}_{x \to y}$ and $\text{NMT}_{y \to x}$.

In this work, we incorporate a Domain-Repair (DR) model in the iterative back-translation process to fully exploit in-domain monolingual data, in which the DR model is used to refine translation sentences given the synthetic bilingual sentences.
The whole framework consists of translation repair and round-trip translation procedures, which are used to generate corresponding training data for NMT and DR models, respectively.
For convenience, we take source-to-target translation ($x \to y$) as an example to explain the usage of our proposed method. 

\paragraph{Translation Repair Stage.} The basic process of back-translation method is to first translate $y^{(t)}$ into $\widehat{x}^{(t)}$ with $\text{NMT}^{0}_{y \to x}$, and then fine-tune $\text{NMT}^{0}_{x \to y}$ on the synthetic parallel data $Y^{*}=\{\widehat{x}^{(t)}, y^{(t)}\}$. 
As the model $\text{NMT}^{0}_{y \to x}$ is not trained on truly in-domain bilingual data, there exists domain mismatch between $\widehat{x}^{(t)}$ and the genuine in-domain sentences $x$.
Given the synthetic parallel data $Y^{*}=\{\widehat{x}^{(t)}, y^{(t)}\}$, we apply the corresponding DR model ($\text{DR}_{ (\widehat{x}, y) \to x}$) to repair errors in translated sentences, e.g. wrong translations of in-domain phrases or domain-inconsistent expressions, 
and then obtain the new synthetic parallel data $\overline{Y}=\{x^{(t)}, y^{(t)}\}$ to train $\text{NMT}_{x \to y}$ initialized with $\text{NMT}^{0}_{x \to y}$. 

\paragraph{Round-Trip Translation Stage.} In order to optimize $\text{DR}_{ (\widehat{x}, y) \to x}$, we use the round-trip translation of monolingual source sentences $X=\{x^{(s)}\}$ to construct the corresponding training data $\widehat{X}=\{\widehat{x}^{(s)}, \widehat{y}^{(s)}, x^{(s)}\}$, where $\widehat{y}^{(s)}$ and $\widehat{x}^{(s)}$ are generated by $\text{NMT}^{0}_{x \to y}$ and $\text{NMT}^{0}_{y \to x}$ respectively ($x^{(s)} \to \widehat{y}^{(s)} \to \widehat{x}^{(s)} $).
In this way, $\text{DR}_{ (\widehat{x}, y) \to x}$ learns to identify mistakes made by $\text{NMT}^{0}_{y \to x}$ and corresponding mapping rules, which helps to better fix the errors in synthetic parallel data.

Similarly, these two stages are also applied in the reverse translation direction to train target-to-source NMT model ($\text{NMT}_{y \to x}$) and corresponding DR model ($\text{DR}_{ (\widehat{y}, x) \to y}$).
As illustrated in Figure \ref{fig:iter-DRBT-overview}, it is natural to extend such a training process to a joint training framework, which alternately carries out the translation repair and round-trip translation procedures to make full use of the advantage of DR models to improve NMT models.

\subsection{Domain-Repair Model} \label{subsec-training-ape}

\begin{figure}[t]
  \centering
\includegraphics[width=0.48\textwidth]{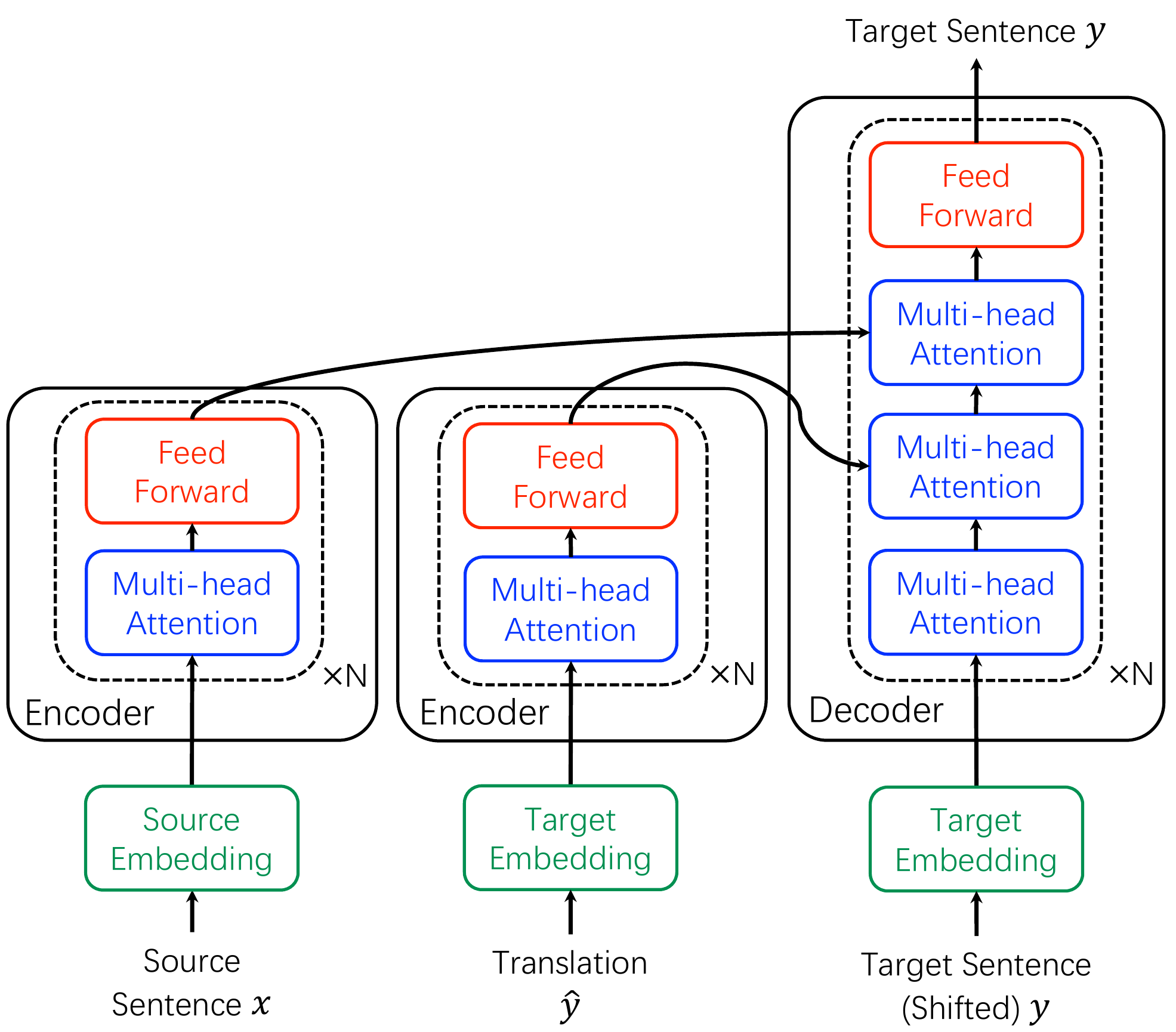}
\caption{The dual-source transformer architecture of the Domain-Repair model ($\text{DR}_{ (\widehat{y}, x) \to y}$). For simplicity, we omit some architecture details such as layer normalization and residual connection.}
\label{fig:dr-model}
\end{figure}

Since the DR model takes the synthetic bilingual sentences as input to produce the in-domain sentences, we parameterize the DR model as a dual-source sequence-to-sequence model. 
As illustrated in Figure \ref{fig:dr-model}, the dual-source transformer model naturally extends the original architecture from \citet{vaswani2017attention} by adding another encoder for translated sentences and stacking an additional multi-head attention component above the multi-head self-attention component.
As usual for the transformer architecture, each block is followed by a skip connection from the previous input and layer normalization.
For simplicity, we omit these architecture details in Figure \ref{fig:dr-model}.

Our proposed framework involves two DR models ($\text{DR}_{ (\widehat{x}, y) \to x}$ and $\text{DR}_{ (\widehat{y}, x) \to y}$), both of which are optimized by maximizing the conditional log likelihood on the training corpus $\widehat{X}=\{\widehat{x}^{(s)}, \widehat{y}^{(s)}, x^{(s)}\}$ and $\widehat{Y}=\{\widehat{x}^{(t)}, \widehat{y}^{(t)},y^{(t)}\}$ built by round-trip translation respectively:
\begin{align}
  \mathcal{L}_1(\theta_1) = \sum_{s=1}^{|\widehat{X}|} \log P( x^{(s)} | \widehat{y}^{(s)}, \widehat{x}^{(s)}; \theta_1) \\
  \mathcal{L}_2(\theta_2) = \sum_{t=1}^{|\widehat{Y}|} \log P( y^{(t)} | \widehat{x}^{(t)}, \widehat{y}^{(t)}; \theta_2) 
\end{align}
where $\theta_1$ and $\theta_2$ denote the model parameters of $\text{DR}_{ (\widehat{x}, y) \to x}$ and $\text{DR}_{ (\widehat{y}, x) \to y}$ respectively.

\begin{algorithm}[t]
    \small
    \SetKw{KwInput}{Input:}
    \KwInput{\textup{pre-trained out-of-domain models} $\textup{NMT}^{0}_{x \to y}$ \textup{and} $\textup{NMT}^{0}_{y \to x}$\textup{, in-domain monolingual sentences} $X=\{x^{(s)}\}$ \textup{and} $Y=\{y^{(t)}\}$\textup{, maximum iteration number} $T$
    } \\
    Use $\text{NMT}^{0}_{x \to y}$ and $\text{NMT}^{0}_{y \to x}$ to perform round-trip translation on $X$ and $Y$ to construct dataset $\widehat{X}=\{\widehat{x}^{(s)}, \widehat{y}^{(s)}, x^{(s)}\}$ and $\widehat{Y}=\{\widehat{x}^{(t)}, \widehat{y}^{(t)},y^{(t)}\}$; \\
    Train $\text{DR}^{0}_{ (\widehat{x}, y) \to x}$ and $\text{DR}^{0}_{ (\widehat{y}, x) \to y}$ with $\widehat{X}$ and $\widehat{Y}$; \\
    $k = 0$; \\
    \For{$k \le T$} {
        \textbf{Translation Repair Stage:} \\
        Use $\text{NMT}^{k}_{x \to y}$ and $\text{NMT}^{k}_{y \to x}$ to build synthetic data $ X^* = \{ x^{(s)}, \widehat{y}^{(s)} \} $ and $ Y^* = \{ \widehat{x}^{(t)}, y^{(t)} \} $  for $X$ and $Y$ respectively; \\
        Use $\text{DR}^{k}_{ (\widehat{y}, x) \to y}$ and $\text{DR}^{k}_{ (\widehat{x}, y) \to x}$ to repair $X^*$ and $ Y^*$ to construct in-domain synthetic data $\overline{X} = \{x^{(s)}, y^{(s)}\}$ and $\overline{Y} = \{x^{(t)}, y^{(t)}\}$; \\
        \textbf{Update NMT Models:} \\
        $\text{NMT}^{k+1}_{x \to y} \leftarrow$ Fine-tune $\text{NMT}^{k}_{x \to y}$ with $\overline{Y}$; \\
        $\text{NMT}^{k+1}_{y \to x} \leftarrow$ Fine-tune $\text{NMT}^{k}_{y \to x}$ with $\overline{X}$; \\
        \textbf{Round-Trip Translation Stage:} \\
        Use $\text{NMT}^{k+1}_{x \to y}$ and $\text{NMT}^{k+1}_{y \to x}$ to perform round-trip translation on $X$ and $Y$ to construct corresponding dataset $\widehat{X}=\{\widehat{x}^{(s)}, \widehat{y}^{(s)}, x^{(s)}\}$ and $\widehat{Y}=\{\widehat{x}^{(t)}, \widehat{y}^{(t)},y^{(t)}\}$; \\
        \textbf{Update DR Models:}\\
        $\text{DR}^{k+1}_{ (\widehat{x}, y) \to x} \leftarrow$ Fine-tune $\text{DR}^{k}_{ (\widehat{x}, y) \to x}$ with $\widehat{X}$;\\
        $\text{DR}^{k+1}_{ (\widehat{y}, x) \to y} \leftarrow$ Fine-tune $\text{DR}^{k}_{ (\widehat{y}, x) \to y}$ with $\widehat{Y}$; \\
        $k = k + 1$
    }
    \caption{Joint Training Algorithm for NMT and DR Models}
    \label{alg:joint-pe-mt}
\end{algorithm}

\subsection{Joint Training Strategy}

We design the iterative training framework to jointly optimize DR and NMT models, as illustrated in Algorithm \ref{alg:joint-pe-mt}.
The whole training framework starts with pre-trained out-of-domain bidirectional NMT models ($\text{NMT}^{0}_{x \to y}$ and $\text{NMT}^{0}_{y \to x}$) and in-domain monolingual data ($X=\{x^{(s)}\}$  and $Y=\{y^{(t)}\}$).
To train initial DR models, we use $\text{NMT}^{0}_{x \to y}$ and $\text{NMT}^{0}_{y \to x}$ to run round-trip translation on $X$ and $Y$ to construct dataset $\widehat{X}=\{\widehat{x}^{(s)}, \widehat{y}^{(s)}, x^{(s)}\}$ and $\widehat{Y}=\{\widehat{x}^{(t)}, \widehat{y}^{(t)},y^{(t)}\}$;

Based on initial NMT and DR models, a joint training process is iteratively carried out to further optimize these models. 
This process consists of translation repair and round-trip translation stages. 
In the translation repair stage, we first adopt NMT models to translate monolingual data, based on which the DR models are used to further re-write the translated sentences as in-domain sentences.
In this way, we can obtain better in-domain synthetic data to further improve NMT models.
Next, in the round-trip translation stage, we perform round-trip translation on monolingual data with enhanced NMT models to re-build training data for DR models. 
The DR models trained on such datasets can better identify mistakes made by latest NMT models ($\text{NMT}^{k+1}_{x \to y}$ and $\text{NMT}^{k+1}_{y \to x}$) and learn corresponding mapping rules, which 
helps to better fix the synthetic parallel data in the next iteration.
Note that we fine-tune the NMT and DR models in each iteration to speed up the whole training process.

\section{Experiments}

\subsection{Setup}

\paragraph{Datasets.} To evaluate the performance of our proposed method, we adopt a multi-domain dataset released by \citet{Koehn2017SixCF}, which is further built as an unaligned monolingual corpus in \citet{Hu2019DomainAO}. 
However, there are two issues in the train/dev/test splits used in \citet{Hu2019DomainAO}. 
First, \citet{ma-etal-2019-domain} and \citet{Dou2020DynamicDS} find that some same sentence pairs exist between the training and test data.
Second, \citet{Hu2019DomainAO} randomly shuffle the bi-text data and split it into halves, which may bring more overlap than in natural monolingual data, i.e., bilingual sentences from a document are probably selected into monolingual data (e.g., one sentence on the source split and its translation on the target split).

To address the impact of the above two issues, we re-collect in-domain monolingual data and test sets in the following steps: 
\begin{itemize}[noitemsep,leftmargin=*,topsep=2pt]
\item Download the XML files from OPUS\footnote{http://opus.nlpl.eu/}, extract parallel corpus from each documents and record the document boundaries.
\item Randomly take some documents as dev/test sets and use the rest as training data.
\item Divide the training set into two parts, where the number of sentences in the two parts is similar. Then the source and target sentences of the first and second halves are chosen as monolingual data, respectively.
\item De-duplicate all overlap sentences within train/dev/test sets.
\end{itemize}
We choose medical (EMEA) and law (JRC-Acquis) domains for our experiments. All the data statistics are illustrated in Table \ref{tab:data-stat}. 


\begin{table}[t]
\footnotesize
  \centering
\begin{tabular}{lrr}
\toprule
\textbf{Domains} & \multicolumn{1}{c}{\textbf{LAW}} & \multicolumn{1}{c}{\textbf{MEDICAL}} \\
\midrule
\midrule
\textbf{\#Bi. } & 377,114  & 328,132  \\
\textbf{\#Mono. (de)} & 187,550  & 171,906  \\
\textbf{\#Mono. (en)} & 189,564  & 156,226  \\
\textbf{\#Dev} & 4,233  & 1,141  \\
\textbf{\#Test} & 4,063  & 1,272  \\
\bottomrule
\end{tabular}%
      \caption{Statistics on bilingual, monolingual, development and test data of medical and law domains.}
    \label{tab:data-stat}%
\end{table}%

\paragraph{Experimental Details.} We implement all NMT models with \textit{Transformer\_base} \cite{vaswani2017attention}. More specifically, the number of layers in the encoder and decoder is set to 6, with 8 attention heads in each layer. 
Each layer in both encoder and decoder has the same dimension of input and output $d_\text{model} = 512$, dimension of feed-forward layer's inner-layer $d_\text{hidden} = 2048$. 
Besides, DR models follow the same setting as the NMT model.

\begin{table*}[t]
\footnotesize
  \centering
\begin{tabular}{lcccccccccc}
\toprule
\multicolumn{1}{c}{\multirow{2}[4]{*}{\textbf{Methods}}} & \multicolumn{2}{c}{\textbf{MED2LAW}} & \multicolumn{2}{c}{\textbf{LAW2MED}} & \multirow{2}[4]{*}{\textbf{Ave.}} & \multicolumn{2}{c}{\textbf{WMT2LAW}} & \multicolumn{2}{c}{\textbf{WMT2MED}} & \multirow{2}[4]{*}{\textbf{Ave.}} \\
\cmidrule{2-5}\cmidrule{7-10}      & \textbf{DE2EN} & \textbf{EN2DE} & \textbf{DE2EN} & \textbf{EN2DE} &       & \textbf{DE2EN} & \textbf{EN2DE} & \textbf{DE2EN} & \textbf{EN2DE} &  \\
\midrule
\midrule
Base & 19.81  & 19.91  & 27.27  & 25.46  & 23.11  & 42.17  & 36.46  & 37.01  & 34.94  & 37.65  \\
Copy  & 20.34  & 20.51  & 29.59  & 27.95  & 24.60  & 42.52  & 36.71  & 37.43  & 37.39  & 38.51  \\
\midrule
BT    & 35.84  & 32.47  & 42.84  & 38.13  & 37.32  & 49.07  & 42.50  & 49.72  & 43.04  & 46.08  \\
DALI-BT  & 36.38  & 33.40  & 44.76  & 39.20  & 38.44  & 49.58  & 42.85  & 50.23  & 43.23  & 46.47  \\
DRBT  & 39.64  & 35.42  & 45.81  & 41.17  & 40.51  & 50.41  & 45.24  & 50.69  & 45.13  & 47.87  \\
\midrule
iter-BT & 40.72  & 33.29  & 45.66  & 40.51  & 40.05  & 49.97  & 44.73  & 51.15  & 45.70  & 47.89  \\
iter-DRBT & \textbf{43.42} & \textbf{37.94} & \textbf{48.69} & \textbf{44.60} & \textbf{43.66} & \textbf{51.15} & \textbf{46.14} & \textbf{51.37} & \textbf{46.04} & \textbf{48.68} \\
\bottomrule
\end{tabular}%
    \caption{BLEU scores(\%) under different settings. The left four columns are results of adapting between two distinct domains, while the right four domains are results of adapting from the general domain (WMT) to specific domains. }
 \label{tab:main-results}%
\end{table*}%

The Adam \cite{kingma2014adam} algorithm is used to update DR and NMT models. 
For training initial NMT and DR models, following the setting of \citet{Hu2019DomainAO}, we set the dropout as 0.1 and the label smoothing coefficient as 0.2.
Besides, we adopt the setting of \textit{Fairseq} \cite{ott2019fairseq} on IWSLT'14 German to English to fine-tune NMT and DR models.
During training, we schedule the learning rate with the inverse square root decay scheme, in which the warm-up step is set as 4000, and the maximum learning rate is set as 1e-3 and 5e-4 for pre-training and fine-tuning, respectively.

For the joint training strategy, we set the maximum iteration number $T$ in Algorithm \ref{alg:joint-pe-mt} as 2 for balancing speed and performance.
In practice, we train our framework on 2 Tesla P100 GPUs for all tasks, and it takes 2 days to finish the whole training.

\paragraph{Methods.} We compare our approach with several baseline methods in our experiment:
\begin{itemize}[noitemsep,leftmargin=*,topsep=2pt]
\item \textbf{Base:} Directly use out-of-domain NMT models to evaluate on in-domain test sets.
\item \textbf{Copy:} Copy the target in-domain monolingual data to the source side as parallel data.
\item \textbf{BT:} Back-translation method, which fine-tunes the out-domain model on synthetic training data generated by a target-to-source out-domain NMT model.
\item \textbf{DALI-BT:} Using word translation instead of back-translation to generate synthetic parallel data. Such data can be mixed with common back-translation for domain adaptation \cite{Hu2019DomainAO}. 
\item \textbf{iter-BT:} Iterative back-translation, which alternatively generates synthetic data and optimizes NMT models at both side \cite{hoang-etal-2018-iterative}. We adopt the same iteration number as iter-DRBT.
\item \textbf{DRBT:} The simplified version of our proposed method, in which we only use the DR model to repair synthetic data once.  
\end{itemize}
All experimental results are evaluated by \textit{SacreBLEU} \cite{Post2018ACF} in terms of case-sensitive tokenized BLEU \cite{Papineni2002BleuAM}.

\subsection{Main Results}

\paragraph{Adapting between Specific Domains.} We verify our approach by adapting NMT models from one distinct domain to another.
As illustrated in the left four columns of Table \ref{tab:main-results}, the unadapted models perform poorly on the out-of-domain test sets. 
Besides, the Copy and BT can improve the performance on target domains, in which the back-translation method achieving more improvements consistently. 
We reproduce \newcite{Hu2019DomainAO}'s work, and their method combined with back-translation (DALI-BT) gains better performance. 
Our proposed method (DRBT) significantly outperforms all previous methods on all four translation tasks, achieving up to average 17.40 and 2.08 BLEU improvements compared to Base and DALI-BT, respectively.
It demonstrates that the DR model effectively repairs the errors occurred by out-of-domain models, improving the performance of unsupervised domain adaptation.

As the back-translation method suffers from low-quality synthetic data, iter-BT is used to improve the quality of synthetic data and achieves 2.73 BLEU improvements on average, but it still has 0.46 BLEU behind DRBT.
This result indicates that the DR model shows a better ability to repair the imperfections of synthetic data. 
The joint training of DR and NMT models (iter-DRBT) can further obtain 3.15 BLEU improvements compared to DRBT.
It also proves that the joint training process helps DR models to better identify mistakes made by the latest NMT models and fix the synthetic parallel data in the following iteration. 

\paragraph{Adapting from General to Specific Domains.} 
We further evaluate our method when adapting a model trained on large amounts of general domain data.
We use out-of-domain models trained on the WMT14 German-English dataset and adapt them to the Medical and Law domains, respectively. 
All results are shown in the right half of Table \ref{tab:main-results}.

These results show a similar pattern as previous experiments, except that the gap between our method and BT/iter-BT is reduced. 
We attribute this reduction to the improvements of general models on in-domain translation. Even so, the iter-DRBT yields the best performance on all test sets, with 11.03 and 0.79 BLEU improvements on average compared to Base and iter-BT, respectively.

\paragraph{Semi-supervised Adaptation.}

\begin{table}[t]
  \centering
  \footnotesize
    \begin{tabular}{ccccc}
    \toprule
    \textbf{\#Para.} & \textbf{BT} & \textbf{DRBT} &  \textbf{iter-DRBT} & \textbf{Sup.} \\
    \midrule
    \midrule
    1K    & 46.03 & 48.98 & 51.30 & 61.56 \\
    5K    & 49.30 & 53.59 & 54.93 & 61.74 \\
    10K   & 51.32 & 54.30 & 56.04 & 62.07 \\
    50K   & 57.99 & 59.29 & 60.03 & 62.81 \\
    \bottomrule
    \end{tabular}%
\caption{BLEU scores(\%) of DRBT and iter-DRBT under semi-supervised scenario with varied size of in-domain parallel data. We also report supervised results with all the in-domain parallel (Sup.) as upper bound.}
  \label{tab:dr-semi-sup}%
\end{table}%

Our method can be easily applied in semi-supervised domain adaptation, with a limited number of in-domain parallel data available. 
The implementation in this setting is to mix the in-domain parallel data with the generated synthetic data for NMT models training.
In addition to the round-translation on monolingual data, we conduct back-translation on parallel data to construct corresponding training data for DR models training.

We conduct experiments on adapting German-to-English NMT models from the Law domain to the Medical domain. 
To assess performance under different scales of in-domain parallel data, we fix the number of monolingual in-domain sentences and vary the number of in-domain parallel sentences in 1K, 5K, 10K, and 50K.
We also report the results of fine-tuning on full in-domain parallel data, including additional in-domain parallel data and monolingual data paired with its original translations, to indicate the upper bound of semi-supervised training. 
All the results are listed in Table \ref{tab:dr-semi-sup}. 
We observe the consistent improvement of our proposed method.
It is worth noting that given 50K in-domain parallel data, the gap between using repaired synthetic data and using the actual parallel data is rapidly reduced from 12.58 to 3.52 BLEU, and further decreased to only 2.78 by joint-training with one more iteration, demonstrating the effectiveness of our method in the semi-supervised scenario.

\subsection{Effect of Joint Training}

\begin{figure}[t]
  \centering
\includegraphics[width=0.47\textwidth]{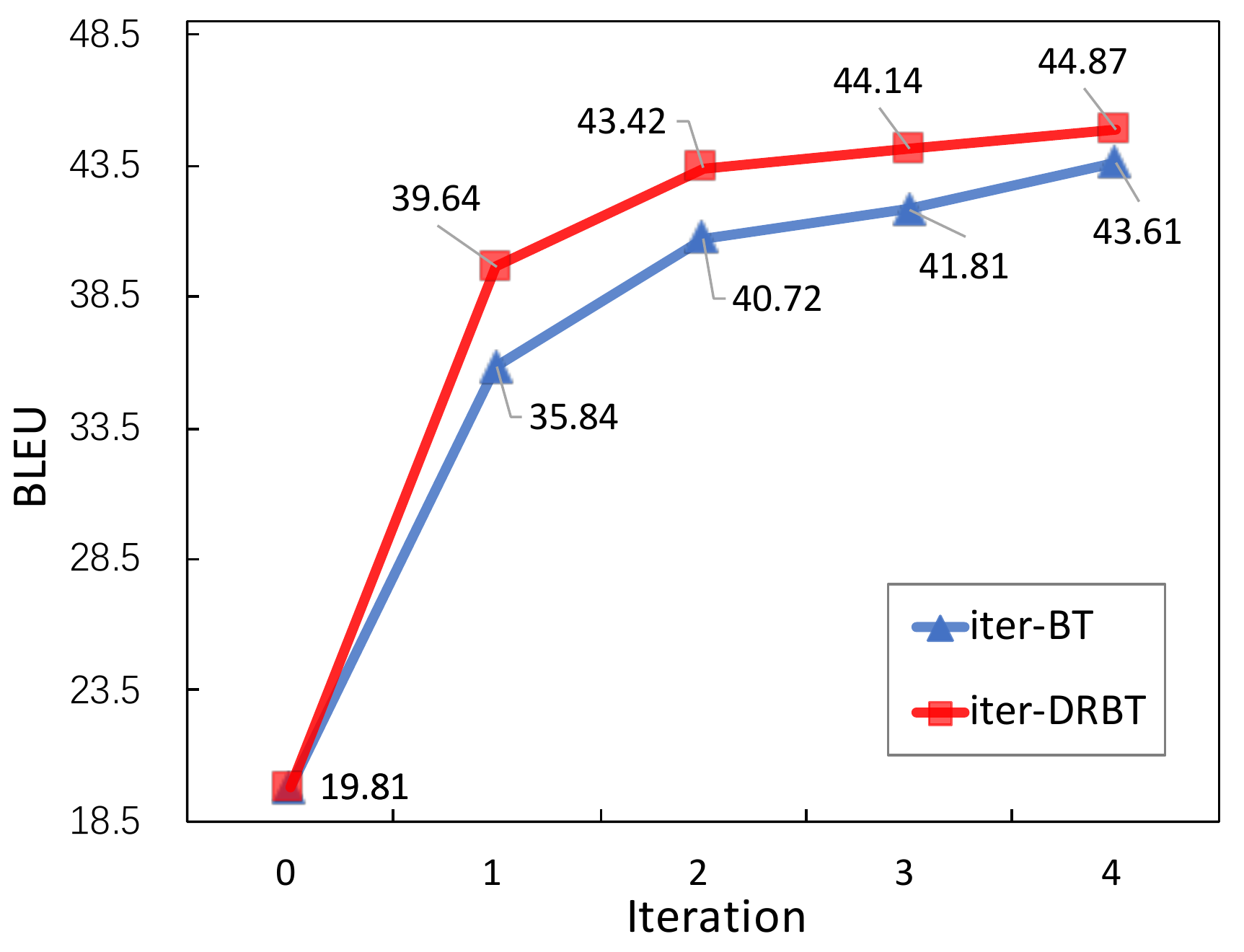}
\caption{BLEU scores(\%) at different iterations of joint training. The model at '0'-th iteration is the unadapted model.}
\label{fig:iter-bt-ape-bt}
\end{figure}

We further investigate the effect of joint training with more iterations. 
Specifically, we conduct experiments on adapting from the Medical domain to the LAW domain from German to English, in which iterative back-translation is used for comparison.

We plot the BLEU curve of these two methods over the number of iterations.
From Figure \ref{fig:iter-bt-ape-bt}, we can observe that our proposed method (iter-DRBT) consistently outperforms iterative back-translation (iter-BT) under the same number of iterations. 
As the number of iterations increases, BLEU improvement achieved by iter-DRBT and iter-BT gradually decreases, but the gap remains.

\subsection{Analysis of Domain Repair Models}
In this section, we mainly discuss how DR models repair the source side of synthetic data to improve its quality. 
Compared to the original back-translation data, we find that the change comes from three main points: an improvement in the overall quality of the source side, an improvement in the accuracy of the in-domain lexical translation, and a closer in-domain style of the source side.

\paragraph{Improvement of Translation Quality.}
 We first assess the change in translation quality at the source side of back-translation data. 
 We report the BLEU changes on all the development sets before and after using the DR model. 
 All the results are listed in Table \ref{tab:ape-analysis-bleu}. 
 We can see that the source side of the back translation data generated by the out domain model is inferior at the initial stage. 
 The DR model significantly improves its quality, which improves the effectiveness of back-translation.

\begin{table}[t]
    \centering
    \footnotesize
    \begin{tabular}{lccc}
    \toprule
          & \textbf{w/o DR} & \textbf{w/ DR} & \textbf{$\Delta$} \\
    \midrule
    \midrule
    \textbf{LAW2MED} & 24.84/26.54 & 36.10/41.06 & 11.2/14.5 \\
    \textbf{MED2LAW} & 18.45/18.46 & 29.80/34.53 & 11.3/16.0 \\
    \textbf{WMT2MED} & 32.62/35.59 & 41.50/46.57 & 8.8/10.8 \\
    \textbf{WMT2LAW} & 34.61/39.87 & 39.48/46.96 & 4.8/7.0 \\
    \bottomrule
    \end{tabular}%
    \caption{BLEU scores(\%) (German/English) on development sets before and after applying DR models.}
    \label{tab:ape-analysis-bleu}%
  \end{table}%


\paragraph{Improvement of Lexical Translation.} 

\begin{figure}[t]
  \centering
  \begin{subfigure}[b]{0.48\textwidth}
  \includegraphics[width=\textwidth]{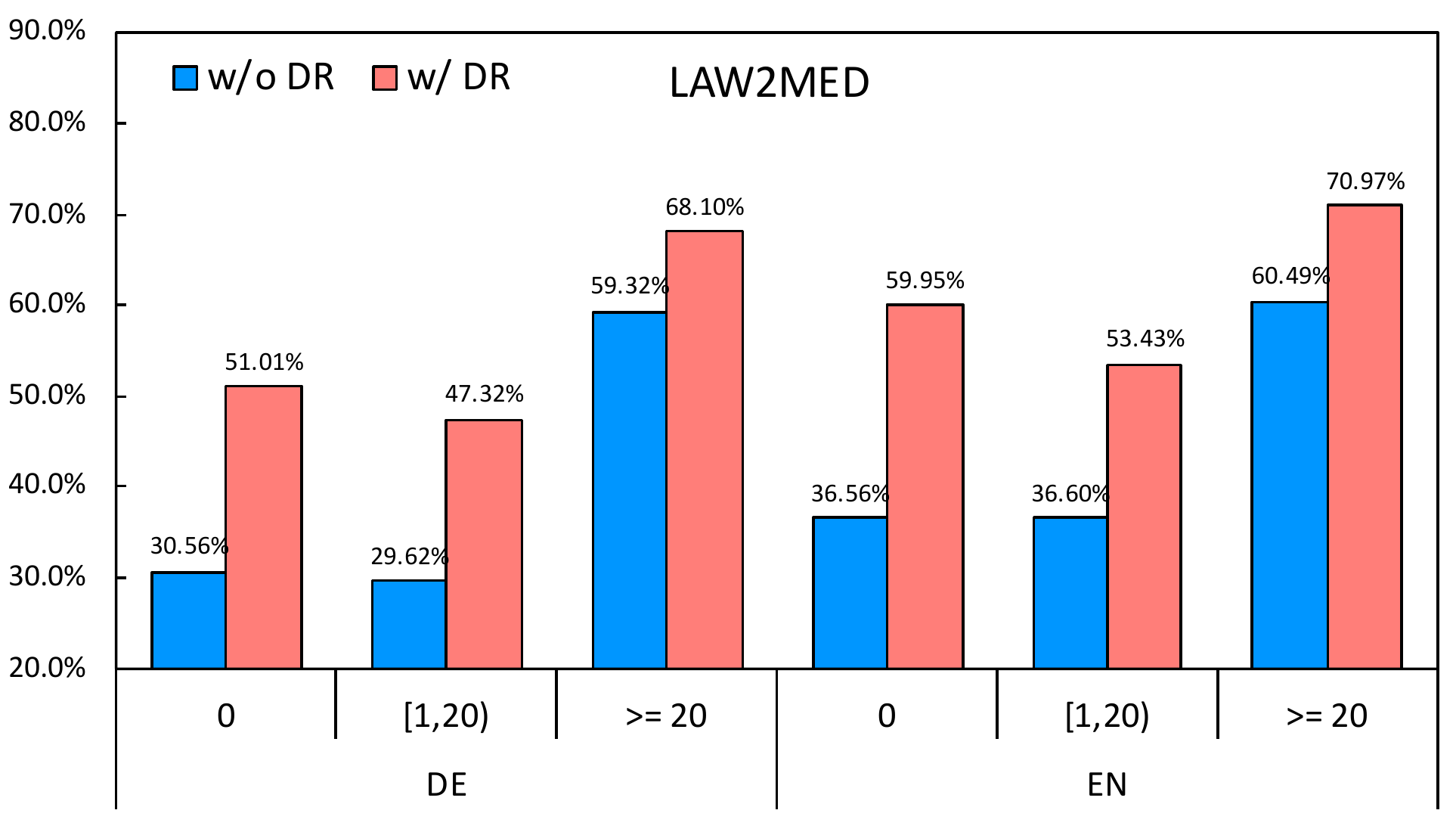}
  \end{subfigure}
    \par\medskip
\begin{subfigure}[b]{0.48\textwidth}
  \includegraphics[width=\textwidth]{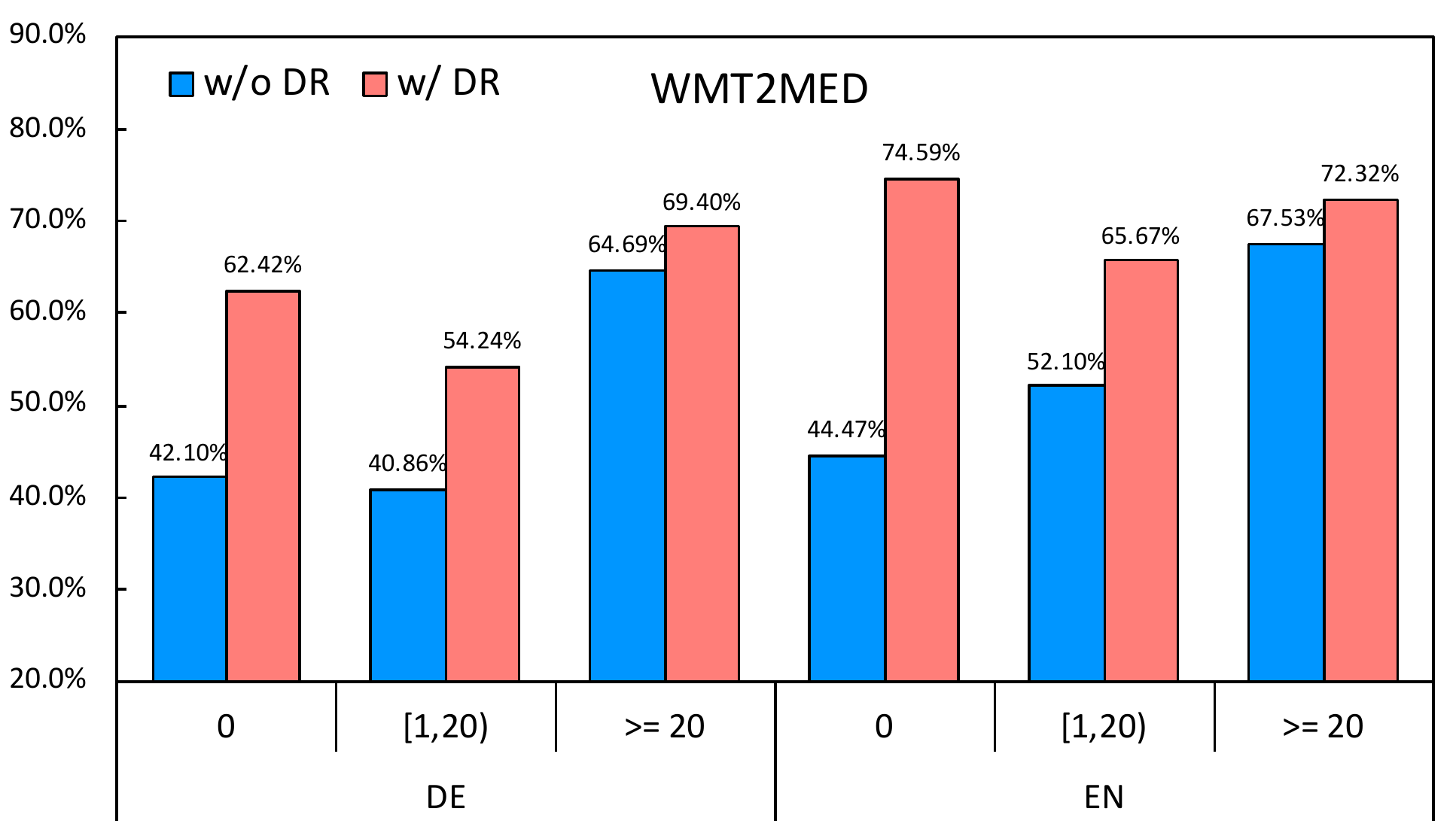}
\end{subfigure}
\caption{F-measures of the word translation on medical development set bucketed by the frequency of words occurring in the out-Of-domain training data.}
\label{fig:ape-wt}
\end{figure}

\begin{table*}[!htbp]
    \centering
    \footnotesize
\begin{tabular}{rp{40.65em}}
\toprule
\textbf{SRC:} & Arzneimittel , deren Plasmaspiegel bei gemeinsamer Anwendung mit Telzir erhöht sein können \\
\textbf{REF:} & Medicinal products whose plasma levels may be increased when \textcolor{red}{\textbf{\uline{co-administered}}} with Telzir \\
\textbf{w/o DR:} & Medicinal products whose plasma ponds may be increased if they are \textcolor{blue}{\textbf{\uwave{commonly used}}} by telzir \\
\textbf{w/ DR:} & Medicinal products whose plasma aspiegel may be increased when \textcolor{red}{\textbf{\uline{co-administered}}} with Telzir \\
\midrule
\textbf{SRC:} & Johanniskraut ( Hypericum perforatum ) Die Serumspiegel von Amprenavir und Ritonavir können durch die gleichzeitige Anwendung von pflanzlichen Zubereitungen mit Johanniskraut ( Hypericum perforatum ) erniedrigt werden . \\
\textbf{REF:} & \textcolor{red}{\textbf{\uline{St John's wort}}} ( Hypericum perforatum ) Serum levels of amprenavir and ritonavir can be reduced by \textcolor{red}{\textbf{\uline{concomitant use of the herbal}}} preparation St John's wort ( Hypericum perforatum ) . \\
\textbf{w/o DR:} & \textcolor{blue}{\textbf{\uwave{Johanniskraut}}} ( Hypericum perforatum ) The serum levels of Amprenavir and Ritonavir can be reduced by \textcolor{blue}{\textbf{\uwave{the simultaneous use of plant}}} preparations with currant ( hypericum perforatum ) . \\
\textbf{w/ DR:} & \textcolor{red}{\textbf{\uline{St. John's wort}}} ( Hypericum perforatum ) Serum levels of amprenavir and ritonavir can be stratified by \textcolor{red}{\textbf{\uline{concomitant use of herbal}}} preparations containing St John's wort ( Hypericum perforatum ) . \\
\bottomrule
\end{tabular}%
    \caption{Cases of sentences that are repaired by DR Model. Inappropriate translations are marked with blue wave lines while corresponding corrections are marked with red underlines.}
    \label{tab:ape-examples}%
  \end{table*}%

We then assess the change in lexical translation at the source side of synthetic data before and after domain repair. Based on the frequency of words that appear in the out-of-domain training data, we allocate target side words of development sets into three buckets ($<1$, $[1, 20)$ and $\ge 20$, which represent zero-shot words, few-shot words, and frequent words, respectively), and compute the word translation f-scores within each bucket. 
We use \textit{compare-mt} \cite{neubig19naacl} to do all the analysis and plot the results in Figure \ref{fig:ape-wt}. 
We can see that the synthetic data repaired by DR models show better word translation in all the buckets. 
It is worth noting that the improvement of word translation f-scores on zero/few-shot ($<20$) words dramatically exceeds that on frequent words, which shows that DR models are especially good at repairing in-domain lexical mistranslations.

\paragraph{Improvement of Domain Consistent Style.}

We further evaluate how can DR models remedy the domain mismatch issue at the source side of back-translated data, including domain inconsistent word selection and language style. 
We evaluate them by observing the perplexity change measured by in-domain and out-of-domain language models before and after being repaired, in which all the language models are trained with \textit{KenLM} \cite{heafield2011kenlm}. 
The out-of-domain language models are trained on out-of-domain training data, while in-domain language models are trained on the original translations of in-domain monolingual data. 
We list all the perplexity scores in Table \ref{tab:dl-lm}. On both MED2LAW and WMT2LAW, we observe a consistent bias of perplexity scores towards in-domain language models, which demonstrates that DR models correct the expression of the source side of synthetic data to be more domain consistent.

\begin{table}[!htbp]
\footnotesize
  \centering
    \begin{tabular}{lcc}
    \toprule
          & Out-of-domain LM & In-domain LM \\
    \midrule
    \midrule
    \multicolumn{3}{c}{\textbf{MED2LAW}} \\
    \midrule
    w/o DR &  15.04/11.16  & 10.93/9.13  \\
    w/ DR &  \textbf{21.17/18.03} $\uparrow\uparrow$ & \textbf{7.27/6.57} $\downarrow\downarrow$ \\
    \midrule
    \multicolumn{3}{c}{\textbf{WMT2LAW}} \\
    \midrule
    w/o DR &  12.29/9.23     & 8.30/6.54 \\
    w/ DR &   \textbf{13.60/9.96} $\uparrow\uparrow$   & \textbf{7.31/5.69} $\downarrow\downarrow$ \\
    \bottomrule
    \end{tabular}%
\caption{Perplexity of synthetic data's source side scores by both in/out domain language models before and after domain repair.}
  \label{tab:dl-lm}%
\end{table}%

\paragraph{Case Study.}

We provide some examples to display how DR models improve the synthetic data.
As shown in Table \ref{tab:ape-examples}, the DR model can reduce some mistranslation, such as correcting the translation of ``Johanniskraut'' into ``St John’s wort'', as well as generating more domain-related expressions, like ``co-administered'' and ``concomitant use of herbal preparations''. 
This shows the ability of domain repair models to improve the quality and domain consistency of synthetic data generated by imperfect out-of-domain NMT models.
\section{Conclusion}

In this paper, we argue that back-translation, the predominant unsupervised domain adaptation method in neural machine translation, suffers from the domain shift, restricting the performance of unsupervised domain adaptation. 
We propose to remedy this mismatch by leveraging a domain repair model that corrects the errors in back-translation sentences.
Then the iterative domain-repaired back-translation framework is designed to make full use of the advantage of the domain repair model.
Experiments on adapting translation models between specific domains and from general domain to specific domains demonstrate the effectiveness of our method, achieving significant improvements over strong back-translation baselines.
 
In the future, we would like to extend our method to enhance the back-translation method in multi-domain settings.

\section{Acknowledgments}

We would like to thank the anonymous reviewers for the helpful comments. This work is supported by National Key R\&D Program of China (2018YFB1403202).

\bibliography{emnlp2020}
\bibliographystyle{acl_natbib}

\end{document}